%% file: root.tex
\documentclass[letterpaper, 10pt, conference]{IEEEtran}      
\usepackage{newtxtext}
\usepackage[italic,defaultmathsizes,symbolgreek]{mathastext}
\usepackage{amssymb, amsmath,mathtools}
\usepackage{nccmath}
\usepackage{hyperref}
\usepackage{cite}
\usepackage{cleveref}
\input{symbols}
\usepackage[protrusion=true,
            expansion=true,
            final]{microtype}
\usepackage{graphicx}
\usepackage{subcaption}
\usepackage{caption}
\usepackage{xcolor}
\usepackage{tabularx}
\usepackage{tikz}
\usepackage{tikz-3dplot}
\usepackage{pgfplots}
\usepgfplotslibrary{external}
\pgfplotsset{compat=1.18}
\pgfmathsetmacro{\plotwidth}{0.8}
\pgfmathsetmacro{\plotheight}{0.7}
\pgfmathsetmacro{\plotheightratio}{0.8}
\pgfmathsetmacro{\subfigratio}{0.9}
\pgfmathsetmacro{\subfigwidth}{0.4}
\usepackage{booktabs}

\usepackage{colortbl}
\definecolor{color1}{HTML}{529DCB}
\definecolor{color2}{HTML}{ECA063}
\definecolor{color3}{HTML}{71BF50}
\definecolor{color4}{HTML}{F3CC4F}
\definecolor{color5}{HTML}{D46934}
\definecolor{color6}{HTML}{A1D8B6}
\definecolor{color7}{HTML}{D2C48E}
\definecolor{color8}{HTML}{F45F40}
\definecolor{color9}{HTML}{F9AE8D}
\definecolor{color10}{HTML}{80B9CE}

\begin{document}
\title{UAV Trajectory Planning with Path Processing}
\author{\IEEEauthorblockN{Zden\v{e}k Bou\v{c}ek, Miroslav Fl\'{i}dr, and Ond\v{r}ej Straka}\\
\IEEEauthorblockA{New Technologies for the Information Society Research Center \& Department of Cybernetics\\
Faculty of Applied Sciences
University of West Bohemia
Pilsen, Czechia\\
Email: zboucek@kky.zcu.cz, flidr@kky.zcu.cz, straka30@kky.zcu.cz}}

\maketitle
\begin{abstract}
This paper examines the influence of initial guesses on trajectory planning for Unmanned Aerial Vehicles (UAVs) formulated in terms of Optimal Control Problem (OCP). The OCP is solved numerically using the Pseudospectral collocation method. Our approach leverages a path identified through Lazy Theta* and incorporates known constraints and a model of the UAV's behavior for the initial guess. Our findings indicate that a suitable initial guess has a beneficial influence on the planned trajectory. They also suggest promising directions for future research.
\end{abstract}


Code: \url{https://github.com/zboucek/TrajectoryPathProcess}

\section{Introduction}
Trajectory planning is critical to autonomous Unmanned Aerial Vehicle (UAV) operations. It determines the vehicle's path, velocity,  orientation, and, in some cases, control signals as functions of time \cite{LaValle2006}. Several distinct approaches exist, including graph-based algorithms for path planning and subsequent trajectory generation, potential field methods, and model-based Optimal Control Problem (OCP) formulations \cite{Betts2010}. Our objective is to solve the trajectory planning using OCP. However, when dealing with nonlinear dynamics, the OCP becomes intractable. To address this challenge, we employed the Chebyshev single and multi-segment pseudospectral method (PSM) \cite{Trefethen2000,Young2019}, a collocation technique that approximates the nonlinear dynamics, integral criterion, and system constraints at collocation points using Chebyshev polynomials. As a result, continuous nonlinear OCP is transcribed into a Nonlinear program (NLP) that can be solved using NLP solvers such as IPOPT \cite{Wachter2006}. Nevertheless, the search for the NLP solution remains challenging. To facilitate the search, we explored leveraging path-planning algorithms and an understanding of the UAV's dynamic model to generate an initial trajectory guess, facilitating the optimization process.

In this paper, we propose an approach to facilitate the solution of NLP by utilizing the graph-based path planning method Lazy Theta* (LT*) to acquire the initial guess. The quality of the initial guess for the NLP directly impacts the convergence of the resulting solution. Traditional methods that employ a simple initial guess, such as a linear or constant path, can lead to poorly converging results.

The papers \cite{Zhou2021} and \cite{Richter2016} both investigate trajectory planning for UAVs. However, their approaches to generating initial guesses for the trajectory planning process differ. The approach taken in~\cite{Zhou2021} involves utilizing an online topological path planning approach to generate a comprehensive set of distinctive paths that guide the optimization process. The approach taken in~\cite{Richter2016} consists of the use of the Rapidly Exploring Random Tree* (RRT*) algorithm to generate an initial route and subsequent construction of a trajectory consisting of a sequence of polynomial spline segments to follow that route. In contrast, our proposed approach is focused on the use of Lazy Theta* (LT*) for the generation of the initial guess.

We chose LT* over RRT* due to its ability to find paths between line-of-sight nodes on a grid map, which can result in more direct paths with fewer waypoints. This can be advantageous for UAVs that can move in any direction, possibly leading to a more efficient trajectory planning process. Based on the LT*-generated path, we construct and test several initial guesses for both state and control trajectories, incorporating UAV constraints, environmental factors, and nonlinear dynamics.


The paper is structured as follows. \Cref{sec:traj-plan} presents the UAV trajectory planning problem defined in terms of OCP, and \Cref{sec:ocp2nlp} describes the process of its transcription to the NLP. \Cref{sec:init-guess} describes the technique for initial guess construction in detail. Furthermore, \Cref{sec:params} presents the details of the implementation and parameters of the trajectory planning problem, along with the process of evaluating solution quality. \Cref{sec:results} presents the results obtained for two distinct environments and discusses these results. Finally, the paper's main points are summarized in \Cref{sec:conclusion}, and future research directions are outlined.
 
\section{Trajectory Planning Problem}
\label{sec:traj-plan}

This section describes the trajectory planning problem in terms of OCP. The OCP includes a nonlinear dynamics model of the UAV, an criterion, and constraints reflecting obstacles, state, and control limitations. 



\subsection{UAV Dynamics Constraints}
\label{sec:model}

The UAV state vector, denoted as $\bdx(t)$, is represented by its position $\position^\localframe$, velocity $\dot{\position}^\localframe$, orientation quaternion $\quaternion$ \cite{Graf2008}, and angular rate $\bdomega$, fully describing the aircraft's motion in 3D space. The state vector is given as 
\begin{equation}
    \bdx(t) = \left[ {\position^\localframe} (t)\tp, {\dot\position^\localframe} (t)\tp, \quaternion (t)\tp, \arate(t)\tp\right]\tp.
\end{equation}
The UAV is actuated by the collective thrust $\force^\bodyframe_T$ and collective torque $\bdtau$. The control vector is given as 
\begin{equation}
    \bdu(t) = \left[\force^\bodyframe_T(t)\tp,\bdtau (t)\tp\right]\tp.
\end{equation}

The UAV equations of motion \cite{Quan2017,Forster2015} are stated below
\begin{align}
\small
	\label{eq:psm-position}
\ddot{\position}^\localframe =& -g \left[0,\,0,\,1\right]\tp + \dfrac{\force^\localframe}{m}, \\
	\label{eq:psm-arate}
\dot{\bdomega} =& -\bdI^{-1}\left(\bdomega \times \bdI \bdomega\right) + \bdI^{-1}\torque, \\
	\label{eq:psm-attitude}
\dot{\quaternion} =& \frac{1}{2} \quatdynmatrix(\quaternion)\tp \bdomega,
\end{align}
where $g$ is the gravity constant, $m$ is the mass of the UAV, $\force^\localframe$ are generalized forces, $\bdI$ is the inertia matrix of the UAV, and  $\quatdynmatrix$ is the quaternion dynamics matrix given as
\begin{equation}
\footnotesize
	\quatdynmatrix(\quaternion) = \begin{bmatrix}
		- q_x & q_w & q_z & -q_y \\
		- q_y & -q_z & q_w & q_x \\
		- q_z & q_y & -q_x & q_w
	\end{bmatrix}.
\end{equation}
The generalized forces $\force^\localframe$ are given as 
\begin{equation}
	\force^\localframe = \quaternion \otimes \left(\collectivethrust^\bodyframe + \forceaerodyn^\bodyframe\right) \otimes \quaternion^* = \bdR(\quaternion)\tp \left(\collectivethrust^\bodyframe + \forceaerodyn^\bodyframe\right),
\end{equation}
where $\quaternion^*$ is the conjugate of $\quaternion$, $\otimes$ is the operator for quaternion product, $\bdR(\quaternion)$ is the body-to-local frame rotation matrix,  and  $\forceaerodyn^\bodyframe$ are aerodynamic forces \cite{Omari2013} given as
\begin{equation}
\forceaerodyn^\bodyframe = -\left(F_{T,z}^{\bodyframe} \right) \cdot \left(\bdK_D \dot\position^\bodyframe\right)\tp,
\end{equation}
where $\bdK_D$ is the lumped drag coefficient matrix and the collective thrust along the $\vecz^\bodyframe$-axis. 

\subsection{Criterion}
\label{sec:opt-crit}
The optimal criterion is an integral component of OCP, as its minimization defines the optimal solution. As the orientation is represented by a quaternion, it is not appropriate to evaluate its deviation by a simple difference. Therefore, the distance between the current orientation $\quaternion(t)$ and the final orientation $\quaternion(t_f)$ is calculated as $\abs{1-\quaternion (t)^T \cdot \quaternion (t_f)}$. To evaluate the quaternions separately, a new symbol $\state_{\state\setminus \quaternion}(t)$ is introduced for the vector, which contains only position, velocity, and angular rate, given as 
\begin{equation}
    \state_{\state\setminus \quaternion}(t) = \left[ \position^\localframe(t)\tp, \dot\position^\localframe (t)\tp, \arate (t)\tp \right]^T.
\end{equation}

The criterion $J$ is defined as
\begin{equation}
    \label{eq:psm-uav-opt-crit}
        J (\bdx(t), \bdu(t), t_0, t_f) = \int_{t_0}^{t_f}\mathcal{L}(\bdx(t), \bdu(t), t, t_f) dt,
\end{equation}
where $t\in [t_0,t_f]$ is the flight time. The function $\mathcal{L}$ is used to quantify the deviations between the UAV's current state $\bdx(t)$ and control $ \bdu(t) $ at any given time $t$ and their respective desired final values $ \bdx (t_f) $ and $\bdu (t_f)$. Function $\mathcal{L}$ can be denoted as
\begin{equation}
    \label{eq:psm-uav-opt-crit-fun}
    \begin{aligned}
        \mathcal{L}&(\bdx(t), \bdu(t), t, t_f) =\\
        =&\left(\state_{\state\setminus \quaternion}(t) - \state_{\state\setminus \quaternion}(t_f)\right)^T \bdQ_{\state\setminus \quaternion} \left(\state_{\state\setminus \quaternion}(t) - \state_{\state\setminus \quaternion}(t_f)\right) \\
         +& \bdQ_{\quaternion} \abs{1-\quaternion (t)^T \cdot \quaternion (t_f)} + \left(\bdu(t) - \control(t_f)\right)^T \bdR \left(\bdu(t) - \control(t_f)\right), \\
    \end{aligned}
\end{equation}
where $\bdQ_{\state\setminus \quaternion}$ and $\bdQ_{\quaternion}$ are positively semidefinite matrices which weight deviations between current state $\bdx(t)$ and $\bdx(t_f)$, and $\bdR$ is positively definite matrix which weights the deviation of control $\bdu(t)$ against $\bdu(t_f)$.

\subsection{Constraints Reflecting Static Obstacles}
\label{sec:obstacles}

The obstacle avoidance system employs constraints that adapt based on the UAV's position and the location of obstacles in the form of columns. The distance from UAV to obstacles is calculated based on the UAV's dimensions and an additional safety margin. The constraints are expressed as 
\begin{equation}
\left(x^{\localframe} (t) - x_{obs}^{\localframe}\right)^2 + \left(y^{\localframe} (t) - y_{obs}^{\localframe}\right)^2 \geq \left(r_{obs} + r_{safe}\right)^2,
\end{equation}
where $x^{\localframe} (t), y^{\localframe} (t)$, $x_{obs}^{\localframe}, y_{obs}^{\localframe}$ are the UAV and obstacle center coordinates, respectively. The radius of the column is denoted as $r_{obs}$. The safety radius around the UAV $r_{safe}$  is calculated as $r_{safe} = \left(l + \frac{d_{p}}{2}\right) \cdot 1.1$, taking into account the arm length $l$ and half the propeller diameter $d_p$ with an additional safety margin of 10\%. To ensure consistency between 2D grid pathfinding and 3D optimal control, the obstacle radius is set to $r_{obs} = \frac{\sqrt{2}}{2} \cdot r_{grid}$, ensuring the entire obstacle is inscribed within the column.

\subsection{State and Control Constraints, and Boundary Conditions}
\label{sec:constraints}

The operation of the UAV is constrained by a set of box constraints on its state and control variables. These constraints were designed based on the laboratory flight space and the UAV specifications. The specific values for these constraints and conditions are provided in \Cref{sec:params}. Additionally, an equality constraint is enforced to maintain the unit quaternion norm, which is necessary to represent a valid orientation.

The initial and final conditions are set for a stabilized UAV state, with the specified position from the grid map. Thus, in the state $\bdx (t_0)$ and $\bdx (t_f)$, the position corresponds to the start and goal, respectively. The quaternion is set to $\quaternion (t_0) = \quaternion (t_f) = \left[1, 0, 0, 0\right]\tp$ and the remaining state elements are set to zero. For the control vector, the thrust is fixed at $F_{T,z}^{\bodyframe} (t_0) = F_{T,z}^{\bodyframe} (t_f) = m\cdot g$ to counter a gravitational force and collective torque is set to zero.  The final time $t_f$ is constrained between 0 and the maximum flight time of the UAV. 

\section{Transcription to Nonlinear Program}
\label{sec:ocp2nlp}

The nonlinear OCP described in \Cref{sec:traj-plan} is intractable. The standard approach to addressing this problem is to employ a numerical method to obtain approximate solutions \cite{Betts2010}. This section presents the PSM \cite{Trefethen2000,Young2019, Ross2012} and its multisegment variant, the pseudospectral elements method (PSEM), and discusses their application, error evaluation, and mesh refinement schemes.

\subsection{Chebyshev Pseudospectral Method}

PSM is a collocation method that approximates problems using high-degree polynomials. It describes the exact situation at collocation points and approximates it elsewhere. We utilize Chebyshev polynomials of the first kind due to their excellent convergence and precision compared to other collocation methods \cite{Boyd2000}. Moreover, the approximation is the most precise at the domain edges due to the placement of the collocation points. This is advantageous because these are the locations where the sharpest changes in state and control trajectories typically occur. Gauss-Lobatto collocation points are used as they include both the roots of the Chebyshev polynomial and the boundary points, allowing direct constraint imposition at boundaries.

Polynomial approximation enables straightforward differentiation and integration of the approximated functions \cite{Trefethen2000}. Derivatives at collocation points are calculated by multiplying the differentiation matrix, while integrals are approximated using the Clenshaw-Curtis quadrature. The number of collocation points determines the placement of these points and the differentiation matrix and integral weights.

Due to its inherent characteristics, the Chebyshev polynomial is only suitable for PSM on the interval $[-1,1]$. This presents a significant challenge, as the OCP is defined on $t\in[t_0,t_f]$. However, a straightforward transformation can facilitate alignment between these intervals \cite{Ross2012}.

To accelerate solution finding and reduce complexity, we also implement PSEM, which approximates the problem across multiple interconnected domains. This approach introduces additional constraints to ensure solution continuity at segment boundaries for time, state, state derivative, and control.

\subsection{Transcription of Optimal Control Problem to Nonlinear Program}

The PSM approximation allows us to transcribe the OCP into an NLP by evaluating the problem only at collocation points and using approximations between them. The criterion is evaluated using integral weights, while state derivatives at collocation points are calculated using a differentiation matrix.

The PSM-approximated OCP with equality and inequality constraints and nonlinear dynamics approximated by an $N$-th degree polynomial can be described as
\begin{align}
\label{eq:ps-nlp-start}
	&\bdz = \left[t_0, t_N, \state_0\tp,\ldots,\state_N\tp,\control_0\tp,\ldots,\control_N\tp\right]\tp,\\
\label{eq:ps-nlp-crit}
	&\min_{\bdz} \sum_{k=0}^N w_k \mathcal{L}\left(\state_k,\control_k,t_k, t_f\right),\\
\label{eq:ps-nlp-diff}
	& \begin{bmatrix}
		0\\\vdots\\0
	\end{bmatrix} = \bdD_N \begin{bmatrix}
		\state_0 \\ \vdots \\ \state_N
	\end{bmatrix} - \begin{bmatrix}
	f\left(\state_0,\control_0,t_0\right) \\ \vdots \\ f\left(\state_N,\control_N,t_N\right)
	\end{bmatrix},\\
\label{eq:ps-nlp-end}
	&g\left(\bdz\right) \leq 0,\,
	h\left(\bdz\right) = 0,\,
	\bdz^-\leq \bdz \leq \bdz^+,
\end{align}
where $\bdz$ includes the boundary time points, state, and control for $N$ collocation points. The integral weights are denoted as $w_k$, and $\bdD_N$ is the differentiation matrix for the $N$-th degree approximation. The approximate criterion is given by \Cref{eq:ps-nlp-crit}.  The dynamics of UAV and its relationship with the approximated derivative is described in \Cref{eq:ps-nlp-diff}, where $f$ includes \Cref{eq:psm-position,eq:psm-arate,eq:psm-attitude}.  Relations \eqref{eq:ps-nlp-end} denote inequality, equality, and box constraints, respectively. Scaling of PSM parameters, namely $t_k$, $w_k$, and $\bdD_N$, is performed according to $t_0$ and $t_f$ using additional equality constraints.

To initialize the NLP solver, it is necessary to input an initial guess, which can have a significant impact on the search for a solution, either by accelerating the process or even identifying a global optimum. The construction of suitable initial guesses is presented in \Cref{sec:init-guess}. The NLP solver is then used to solve the NLP, with trajectories constructed by fitting the extracted state and control collocation points with an appropriate degree polynomial.


\subsection{Solution Error and Mesh Refinement Schemes}
\label{sec:error-and-scheme}

Solving complex problems using PSM and PSEM typically requires an iterative process to ensure sufficient solution accuracy. This section describes the evaluation of the solution and the iterative mesh refinement method used to increase accuracy.
To evaluate the accuracy of the PSM, we compute the discretization error $\epsilon_d (t)$ as the difference between the derivative of the state polynomial and the nonlinear dynamics function
\vspace{-0cm}
\begin{equation}
\epsilon_d (t)  = \dfrac{d p_{\bdx}(t)}{dt}  - f(p(\bdx(t)), p(\bdu(t)),t),
\end{equation}
where $p(\bdx(t))$ and $p(\bdu(t))$ are polynomial approximations of state and control trajectories, respectively.
The absolute discretization error $\epsilon_{a,i}$ between collocation points is
\vspace{-0.1cm}
\begin{equation}
\label{eq:abs-error}
\epsilon_{a,i} = \int_{t_i}^{t_{i+1}} \abs{\epsilon_d(t)} dt,
\end{equation}
evaluated using composite Simpson's rule with $N_{simp} = 10$ points. The relative error $\epsilon_{r,i}$ is calculated as
\vspace{-0.1cm}
\begin{equation}
\small
\epsilon_{r,i} = \dfrac{\epsilon_{a,i}}{\dfrac{1}{N_{simp}}\sum_{t_l=t_i}^{t_{i+1}}{p_{\bdx}}(t_l)}.
\end{equation}
We define maximum absolute and relative errors as
\vspace{-0.1cm}
\begin{equation}
\epsilon_{a_{\max},i} = \max\epsilon_{a,i},\quad \epsilon_{r_{\max},i} = \max\epsilon_{r,i}.
\end{equation}
The PSM iteratively increases the polynomial degree until the error tolerance is reached, determined by
\vspace{-0.1cm}
\begin{equation}
P_k = \max\left(\lceil{\ln(N_k, \epsilon_{a_{\max}}/\epsilon)}\rceil, 3\right),
\end{equation}
where $P_k$ is the increase in the number of collocation points, $N_k$ the original number of points, and $\epsilon$ the error tolerance.
The PSEM scheme differs by allowing segment division based on relative error. If $\epsilon_{r_{\max}}$ exceeds a tolerance value, the segment is divided at the point with the highest relative deflection
\vspace{-0.1cm}
\begin{equation}
\Delta\epsilon_{r_{\max},i} = \epsilon_{r_{\max},i+1} - \epsilon_{r_{\max},i}.
\end{equation}
The refined mesh is determined, and polynomial fitting propagates the solution to the new collocation points. This refined solution serves as the initial guess for the next iteration in the search for the optimal solution.

\section{Initial Guess through Graph-Based
Path Planning}
\label{sec:init-guess}

While simple linear interpolation between boundary conditions is a common approach for initial guesses, it can be ineffective with nonlinear constraints or nonconvex obstacles. We propose a set of more complex initial guesses with varying degrees of influence on the trajectory, as summarized in \Cref{tab:init-guess-levels}.

\begin{table}
    \caption{Summary of initial guess construction for state and control (separated by the line)}
    \label{tab:init-guess-levels}
    \resizebox*{\linewidth}{!}{
    \begin{tabular}{lll}
        \hline
        \textbf{Component} & \textbf{Method} & \textbf{Purpose} \\ \hline
        Simple & Straight line interpolation & Basic path planning \\
        Position & Spline & Smooth path following \\
        Velocity & Differentiation of position & Smooth velocity profile \\ 
        Orientation & Quaternion curve & Align\,with\,forces \\ 
        Angular rate & Quaternion derivative & Orientation changes \\ \hline
        Thrust & Rotation of force & Translation and orientation \\ 
        Torque & Dynamic\,equation & Desired angular motion \\ \hline
    \end{tabular}}
\end{table}

Our initial guess leverages the LT* graph-based path planning algorithm \cite{Nash2010}, which extends A* to identify direct paths between visible grid map nodes. Time parametrization follows known velocity constraints, providing an optimistic time frame. Other state and control guesses are derived from the UAV dynamics (Table \ref{tab:init-guess-levels}).

\subsection{Time Parameterization}
\label{sec:time}

The LT* algorithm assigns waypoints $\bdS = [\bds_0,\bds_1,\ldots,\bds_{M}]$ along the path as locations for trajectory segments, where $\bds_0 = \bds_{start} = \bdr^\localframe_0$ and $\bds_{M} = \bds_{goal} = \bdr^\localframe_f$. The distances between successive waypoints are calculated as $\Delta \bds_i = \abs{\bds_{i+1} - \bds_{i}}$

For 2D paths, the $z$-axis element is linearly interpolated based on boundary conditions
\begin{equation}
{z^\localframe}\left(t\right) = z^\localframe_0 + \dfrac{t - t_0}{t_f - t_0} \cdot \left(z^\localframe_f - z^\localframe_0\right).
\end{equation}
The time required for the UAV to travel between waypoints, considering maximum velocity constraints, is computed as
\begin{equation}
\Delta t_f^{(i)} = \max \left(\frac{\Delta \bds_i}{\dot{\bdr}^\localframe_{\max}}\right), i = 0,\ldots,M-1.
\end{equation}
The total optimistic time is the sum of these intervals $t_f = \sum_{i=0}^{M-1} \Delta t_f^{(i)}$.
The time grid is generated based on the number of collocation points per segment. For multi-segment cases, the grid is constructed sequentially for each segment span. PSEM can also be initialized with a single-segment initial guess.

\subsection{State and Control Initial Guess}
\label{sec:state-ctrl-guess}

We propose a hierarchical approach (as presented in \Cref{tab:init-guess-levels}) to generate initial guesses for state and control variables, building from simple interpolations to more complex dynamics-based guesses based on relations in \Cref{sec:model}:
\begin{itemize}
\item Simple: A straight linear interpolation between initial and final states and controls, with $t_f = \frac{1}{2}(t_{f_{\max}} + t_{f_{\min}})$.
\item Position: LT* path waypoints are fitted with a cubic spline, parameterized by the established time grid.

\item Velocity: Differentiation of the position trajectory
\begin{equation}
    \dot\position^\localframe =  \tfrac{d \position^\localframe}{dt}
\end{equation}

\item Orientation: Using expected force over time
\begin{equation}
    \bdF_r^\localframe = m \cdot (\ddot{\position}^\localframe + g \vecz^\localframe)
\end{equation}
Quaternion curve is generated using
\begin{equation}
\footnotesize
	\quaternion = \dfrac{1}{\sqrt{2(1+ \dfrac{\forceBdes}{\norm{\forceBdes}} \cdot \dfrac{\forceLdes}{\norm{\forceLdes}})}}\begin{bmatrix}
		1+\dfrac{\forceBdes}{\norm{\forceBdes}} \cdot \dfrac{\forceLdes}{\norm{\forceLdes}}\\
		\dfrac{\forceBdes}{\norm{\forceBdes}} \times \dfrac{\forceLdes}{\norm{\forceLdes}}
    \end{bmatrix}
\end{equation}

\item Angular Rate: Derived from quaternion trajectory
\begin{equation}
\arate = 2 \quatdynmatrix(\quaternion) \dot\quaternion = -2\quatdynmatrix(\dot\quaternion)\quaternion
\end{equation}

\item Thrust: Obtained through quaternion rotation of expected force to body frame
\begin{equation}
    \forceBdes = \quaternion^{-1} \otimes \forceLdes \otimes \quaternion, \bdF^B_{T,z} = \bdF^B_{r,z}
\end{equation}

\item Torque: Based on rotational dynamics equation
\begin{equation}
    \torque= -\bdI\dot{\bdomega} + (\bdomega \times \bdI \bdomega)
\end{equation}
\end{itemize}
This hierarchical approach provides increasingly refined initial guesses, potentially improving the efficiency of subsequent optimization processes. Each level builds upon the previous, incorporating more of the UAV's dynamics and constraints.

\section{Parameters and Implementation}
\label{sec:params}

This section presents the values for the parameters of the trajectory planning problem and describes the software and hardware sources employed in the solution acquisition process. This section aims to provide further insight into the solved problem and assist in understanding the results presented in \Cref{sec:results}.

The parameters of the UAV are determined by the specifications of the nano UAV Crazyflie\footnote{Crazyflie -- \url{https://www.bitcraze.io/products/crazyflie-2-1/}}. The specific values of the UAV are set based on \cite{Forster2015} and are presented in \Cref{tab:uav_params}. The lumped drag coefficient matrix is set as 
\begin{equation}
\footnotesize
    \bdK_D = -1 \cdot 10^{-7} \cdot \begin{bmatrix}
        10.2506 & 0.3177 & 0.4332 \\
        0.3177 & 10.2506 & 0.4332 \\
        7.7050 & 7.7050 & 7.5530 \\
    \end{bmatrix}.
\end{equation}


\begin{table}[ht!]
    \centering
    \caption{Crazyflie UAV Model Parameters and Constraints}
    \label{tab:uav_params}
    \begin{tabular}{l l l}
    \hline
    Parameter & Symbol & Value \\
    \hline
    Gravitational acceleration & $g$ & 9.81305 m/s$^2$ \\
    UAV mass & $m$ & 0.032 kg \\
    Arm length & $l$ & 0.0397 m \\
    Moment of inertia, $\vecx^\bodyframe$-axis & $I_x$ & $6.410179 \cdot 10^{-6}$ kg$\cdot$m$^2$ \\
    Moment of inertia, $\vecy^\bodyframe$-axis & $I_y$ & $6.410179 \cdot 10^{-6}$ kg$\cdot$m$^2$ \\
    Moment of inertia, $\vecz^\bodyframe$-axis & $I_z$ & $9.860228 \cdot 10^{-6}$ kg$\cdot$m$^2$ \\
    Propeller diameter & $d_p$ & 0.051 m \\
    maximum flight time & $t_f$ & 3 min\\
    \hline
    \end{tabular}
\end{table}

The state box constraints have the following specific values
\begin{gather}
\small
    \left[-2,-2,0\right]\tp \leq \position^\localframe (t) \leq \left[2,2,2\right]\tp,\,
    -2 \leq \dot\bdr^{\localframe} (t) \leq 2,\\
    0 \leq q_w (t) \leq 1,\,
    -1 \leq \bdq_r (t) \leq 1,\,
    -\infty \leq \arate (t) \leq \infty,
\end{gather}
and the control box constraints are as follows
\begin{equation}
\small
    0 \leq F_{T,z}^{\bodyframe} (t) \leq 0.6, \torque^- \leq \torque (t) \leq \torque^+,
\end{equation}
where $\torque^+ = -\torque^- = \left[5.955,5.955,1.82063\right]\tp\cdot10^{-3}$.

The weight matrices in \Cref{eq:psm-uav-opt-crit-fun} are set as
\begin{equation}
\small
    \begin{aligned}
    \bdQ_{\bdx\setminus \quaternion} =& \text{diag}\left(1, 1, 1, 1, 1, 1, 0.328, 0.328, 0.328\right),\,\bdQ_{\quaternion} = 100,\\
    \bdR =& \text{diag}\left(1.223 \cdot 10^{1}, 2.820\cdot 10^{4}, 2.820 \cdot 10^{4}, 3.017 \cdot 10^{5}\right).
    \end{aligned}
\end{equation}

The trajectory planning algorithm, which incorporates PSM and PSEM, was developed in Python using Pyomo \cite{Bynum2021} for optimization modeling. The IPOPT solver \cite{Wachter2006} was employed to solve the resulting large-scale nonlinear programming problems. The UAV's dynamic model from \Cref{sec:model} was incorporated using SymPy for symbolic computations.

To ensure consistency, the algorithm was executed on the Czech National Grid Infrastructure MetaCentrum\footnote{MetaCentrum -- \url{https://metavo.metacentrum.cz/en/about/index.html}}, with a minimum CPU computational power of 8.0 SPECfp2017\footnote{SPECfp2017 norm -- \url{https://www.spec.org/cpu2017/Docs/}}. All computations were performed on AMD Epyc processors, with 16 CPUs and 50 GB of RAM allocated per job. A custom Apptainer\footnote{Apptainer -- \url{https://apptainer.org/}} image with Miniconda was created to ensure consistent execution across different machines, simplifying the deployment of IPOPT and other dependencies.

\section{Results}
\label{sec:results}

\begin{figure}
    \centering
    \begin{subfigure}[t]{0.3\linewidth}
        \centering
        \includegraphics[width=\linewidth]{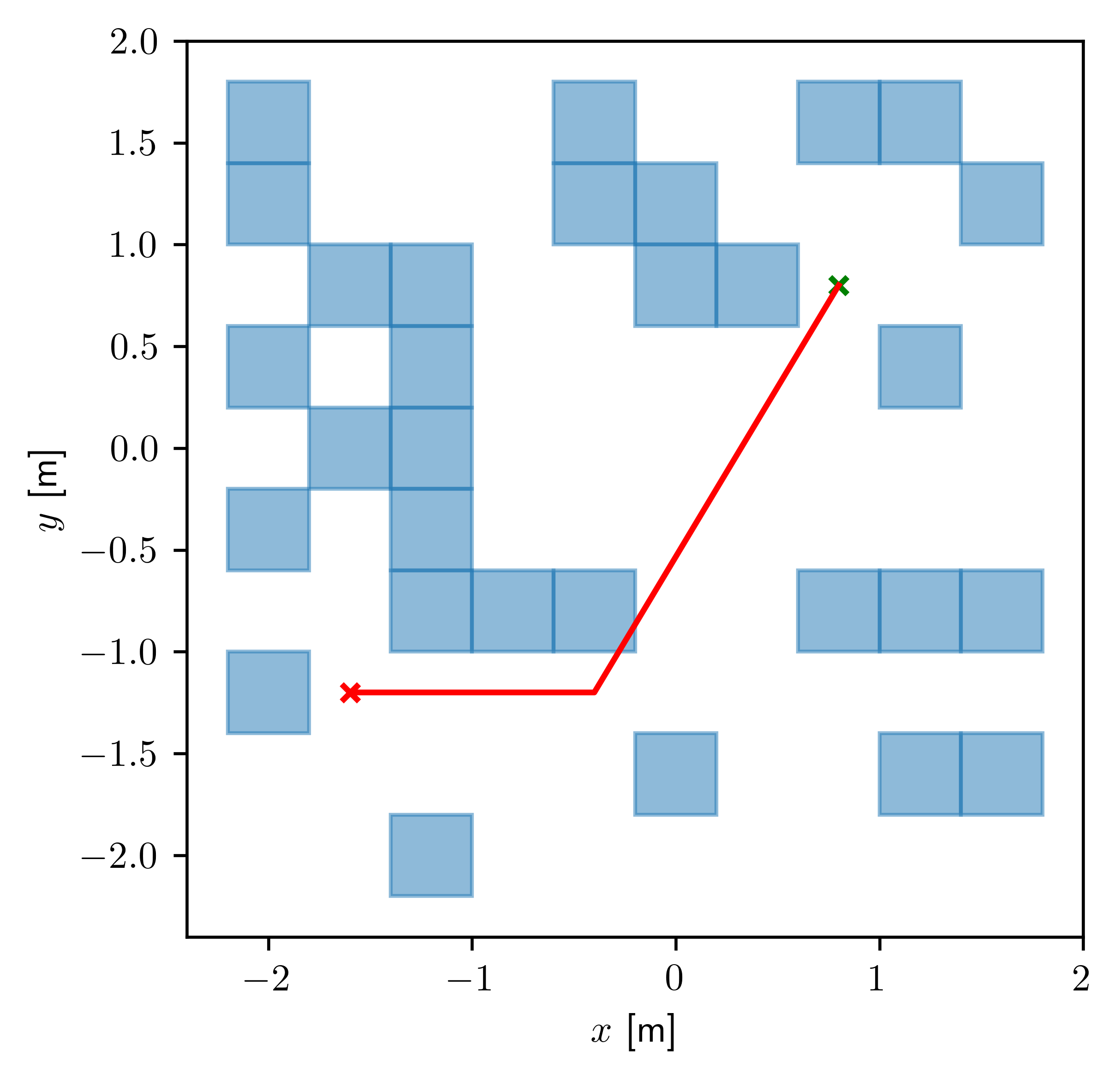}
        \caption{LT* path}
        \label{fig:path-example-rand-columns}
    \end{subfigure}
    \hfill
    \begin{subfigure}[t]{0.35\linewidth}
        \centering
        \includegraphics[width=\linewidth]{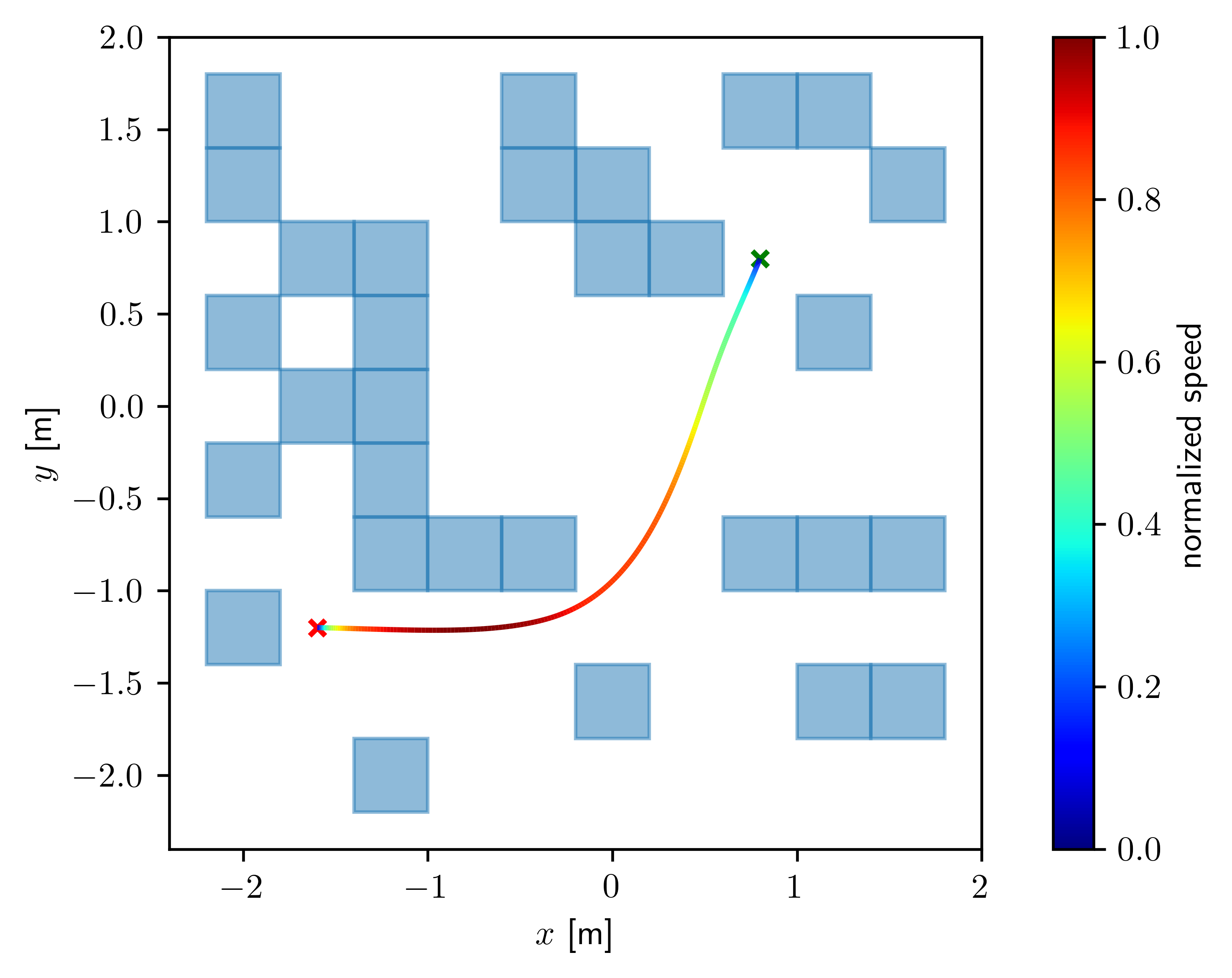}
        \caption{Trajectory}
        \label{fig:traj-example-random-columns-state-2dtraj}
    \end{subfigure}
    \hfill
    \begin{subfigure}[t]{.28\linewidth}
        \centering
        \includegraphics[width=\linewidth]{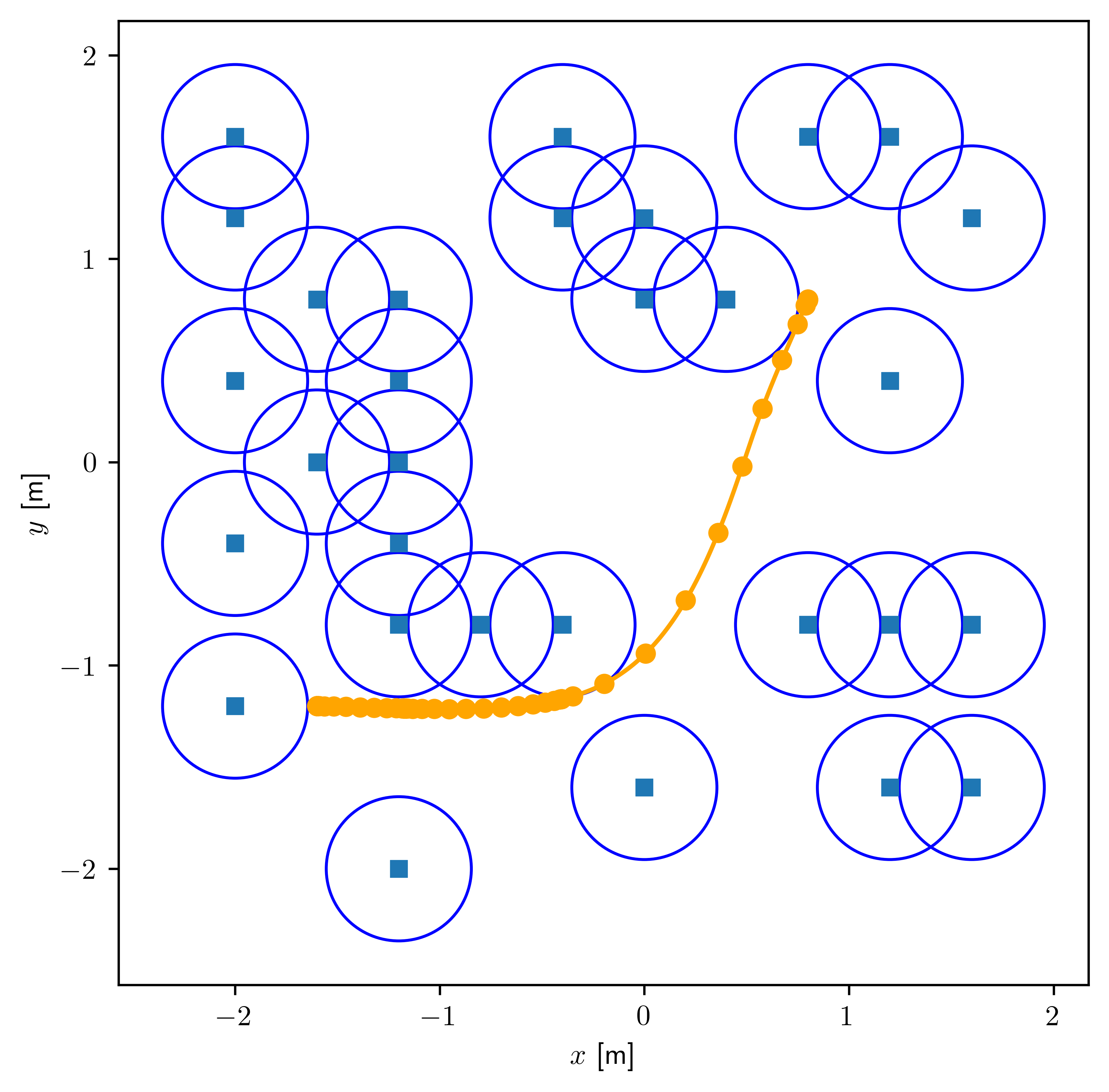}
        \caption{Collocation points}
        \label{fig:traj-example-random-columns-state-2dposition}
    \end{subfigure}
    \caption{Path and trajectory in the environment}
    \label{fig:traj-plan-example-random-columns-xy}
\end{figure}

This section presents the results for two critical scenarios: a two-column obstacle and a random columns environment. We focus on these scenarios as they provide the most informative insights into the algorithm's performance while adhering to space constraints. 
Figure \ref{fig:path-example-rand-columns} illustrates an LT* path example for a random columns environment. Figure \ref{fig:traj-example-random-columns-state-2dtraj} shows an optimal trajectory with a speed profile, while Figure \ref{fig:traj-example-random-columns-state-2dposition} depicts the position at the collocation points and obstacle constraints.

The results are analyzed based on several criteria:
\begin{itemize}
\item \textit{Init. Level}: The initial guess complexity level (\Cref{sec:init-guess}).
\item \textit{Constr.}: Inclusion of boundary conditions and state/control limits in the initial guess.
\item \textit{Method}: PSM (single segment) or PSEM (multi-segment).
\item \textit{Iter.}: Number of iterations for convergence.
\item \textit{Criterion}: \Cref{eq:psm-uav-opt-crit}, approximated by Clenshaw-Curtis quadrature.
\item \textit{Absolute Error}: Maximum error across the trajectory based on \Cref{eq:abs-error}.
\item \textit{Sum Viol.}: Total constraint violation.
\item \textit{Obstacle Viol.}: Obstacle interference.
\item \textit{Total Time}: Computation time on the MetaCentrum grid in seconds.
\end{itemize}
It is crucial to acknowledge that, as PSM and PSEM are collocation methods, constraints are only evaluated at collocation points, which may occasionally result in obstacle encounters. The discretization error, which is employed as a stopping metric for the trajectory planning algorithm, is set with a threshold of $\epsilon_{a_{\max}} = 10^{-2}$. The search is limited to 10 iterations or 5 hours if the aforementioned threshold is not reached. In some cases, this can result in local minima that fail to satisfy all constraints.
In the following tables, the best values are marked in green, the worst in orange, and the 90th and 10th percentiles are marked in light red and blue, respectively.

\subsection{Two Obstacles Scenario}

\begin{figure}\captionsetup[subfigure]{font=footnotesize}
    \centering
    \begin{subfigure}[t]{0.3\linewidth}
        \centering
        \includegraphics[width=\linewidth]{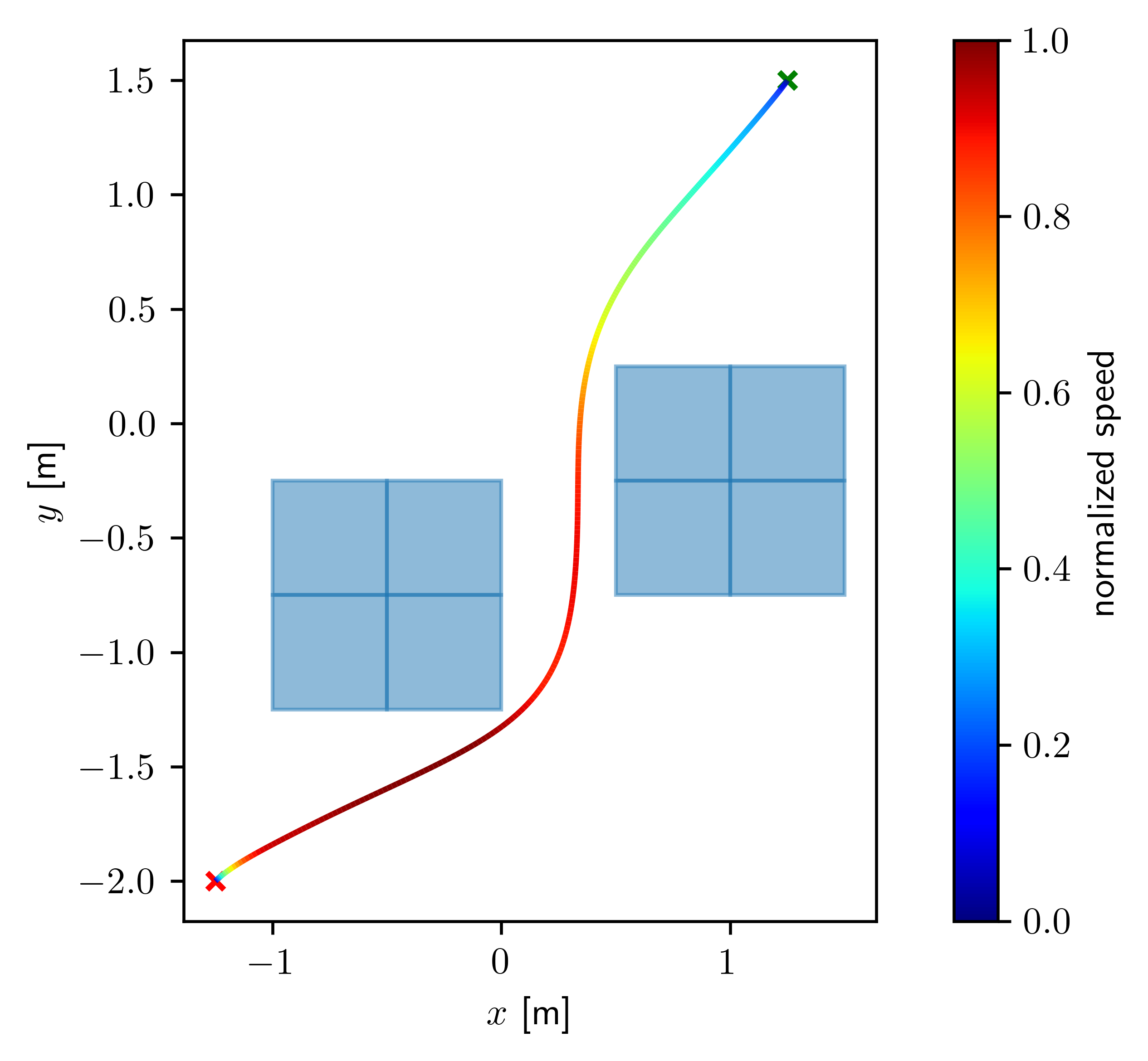}
        \caption{PSM single-seg. orientation init. and forced constraints}
        \label{fig:traj-simple2-orientation-psm}
    \end{subfigure}
    \hfill
    \begin{subfigure}[t]{0.3\linewidth}
        \centering
        \includegraphics[width=\linewidth]{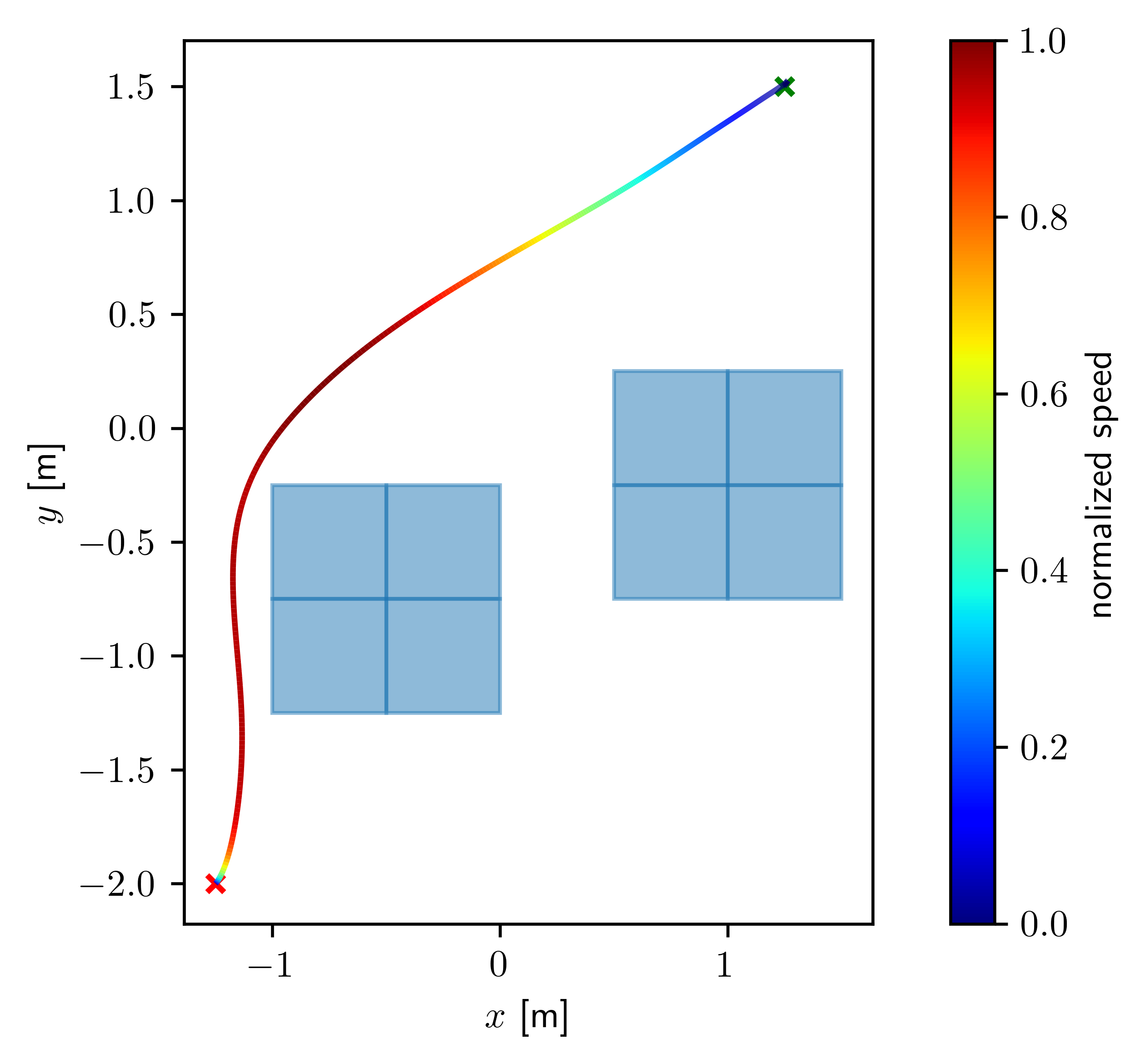}
        \caption{PSEM single-seg. a. rate with control init. and forced constraints}
        \label{fig:traj-simple2-arate-ctrl}
    \end{subfigure}
    \hfill
    \begin{subfigure}[t]{0.3\linewidth}
        \centering
        \includegraphics[width=\linewidth]{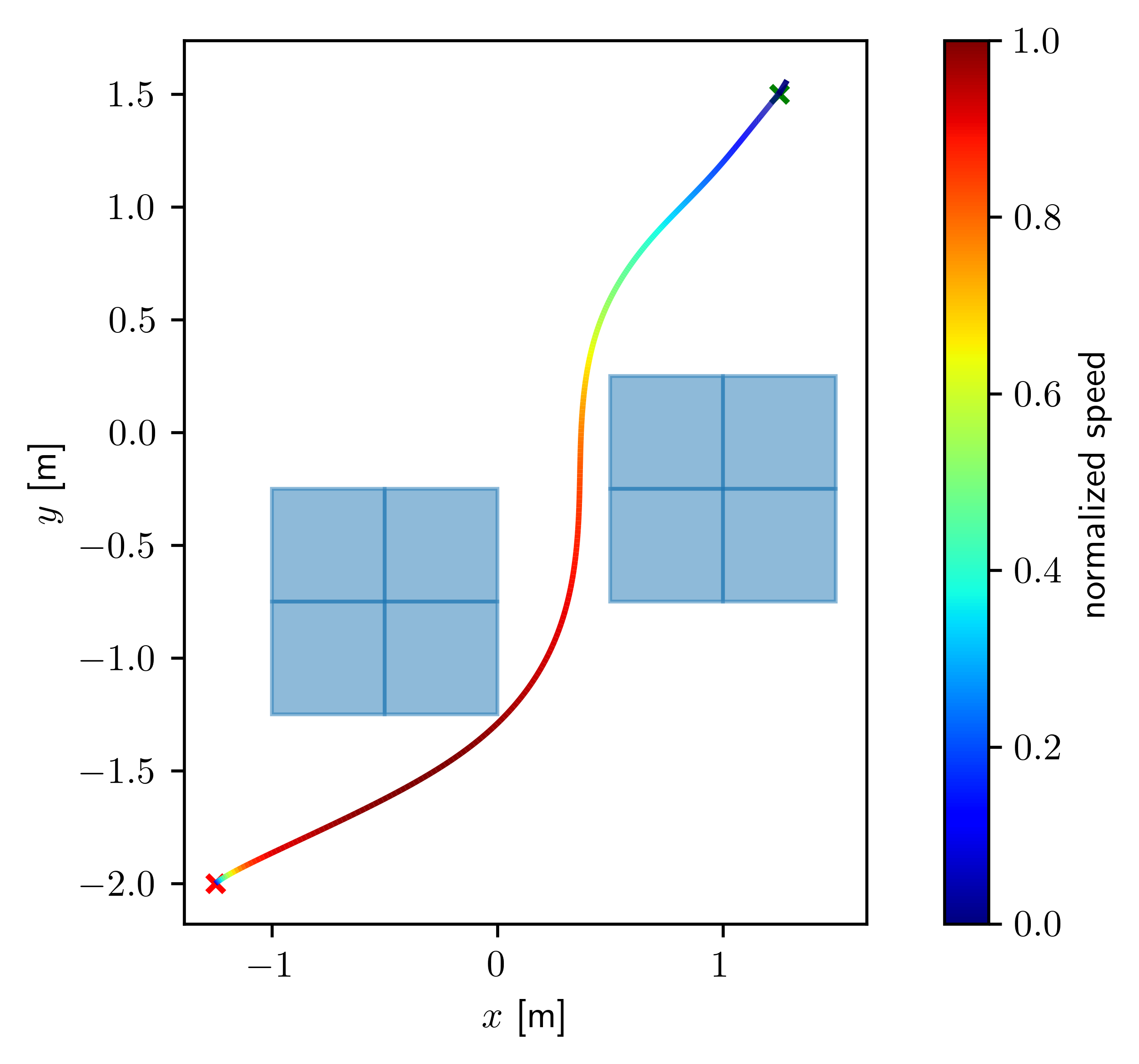}
        \caption{PSEM multi-seg. orientation init. and forced constraints}
        \label{fig:traj-simple2-orientation-psem}
    \end{subfigure}
    \caption{PSM and PSEM trajectory planning for two obstacles}
    \label{fig:traj-plan-simple2}
\end{figure}

In the two-obstacle scenario (\Cref{fig:traj-plan-simple2}), most trajectories followed the LT* path, with no trajectories found without constraint enforcement. PSEM with single-segment angular rate and control initialization (\Cref{fig:traj-simple2-arate-ctrl}) found a unique path in the shortest time, while simple initialization consistently failed.

\Cref{tab:traj-plan-eval-simple2-single} shows position initialization required the least iterations but resulted in the largest violations. PSM with orientation initial guess (\Cref{fig:traj-simple2-orientation-psm}) achieved the best criterion and absolute error. Multi-segment initialization (\Cref{tab:traj-plan-eval-simple2-multi}) yielded slightly lower optimality values but often increased computation time.

\input{tables/table_traj_plan_eval_simple2_multi}
\input{tables/table_traj_plan_eval_simple2_single}

\subsection{Random Columns Scenario}

\begin{figure}\captionsetup[subfigure]{font=footnotesize}
    \centering
    \begin{subfigure}[t]{0.45\linewidth}
        \centering
        \includegraphics[width=\linewidth]{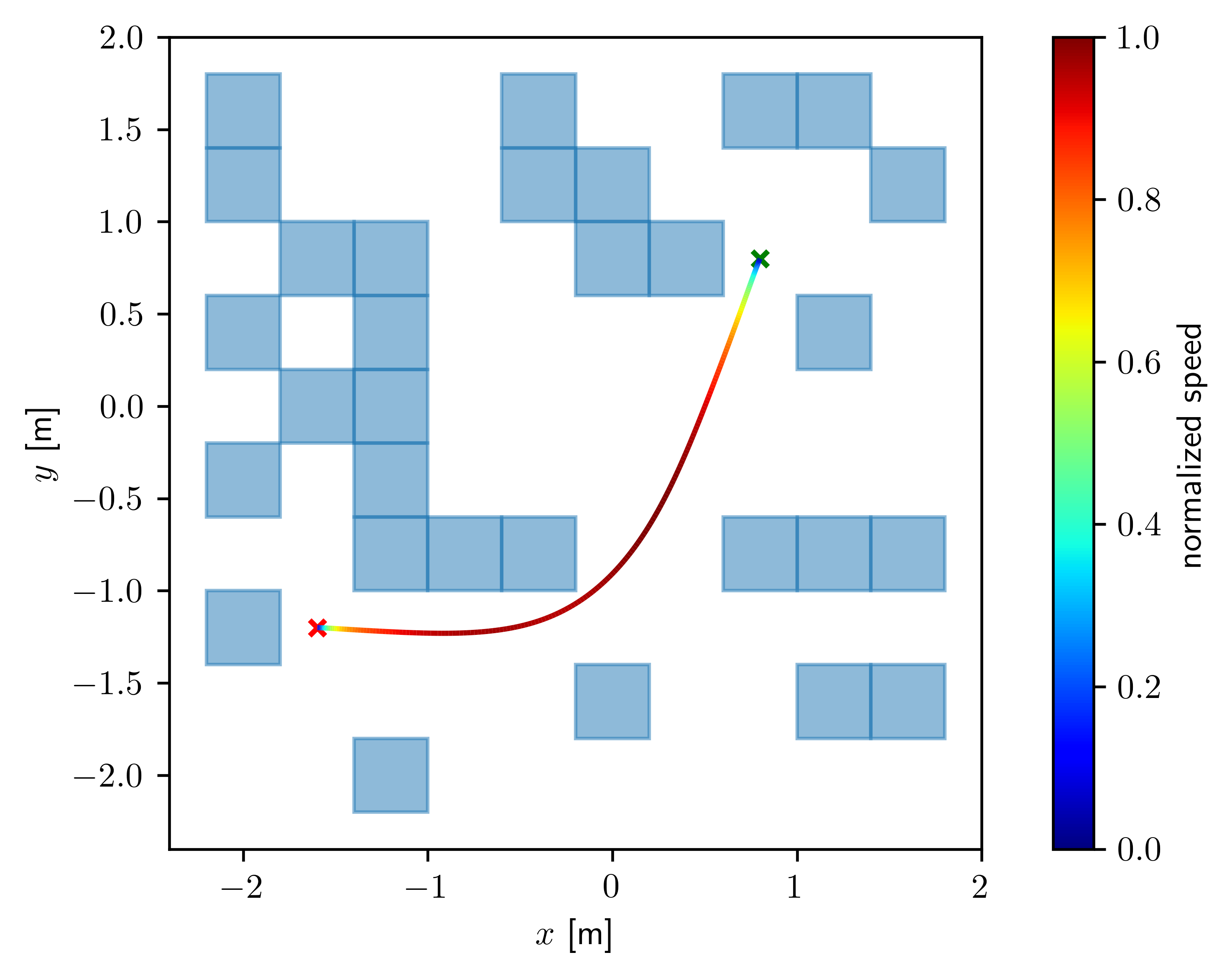}
        \caption{single-segment simple init.}
        \label{fig:traj-random-columns-none-single}
    \end{subfigure}
    \hfill
    \begin{subfigure}[t]{0.45\linewidth}
        \centering
        \includegraphics[width=\linewidth]{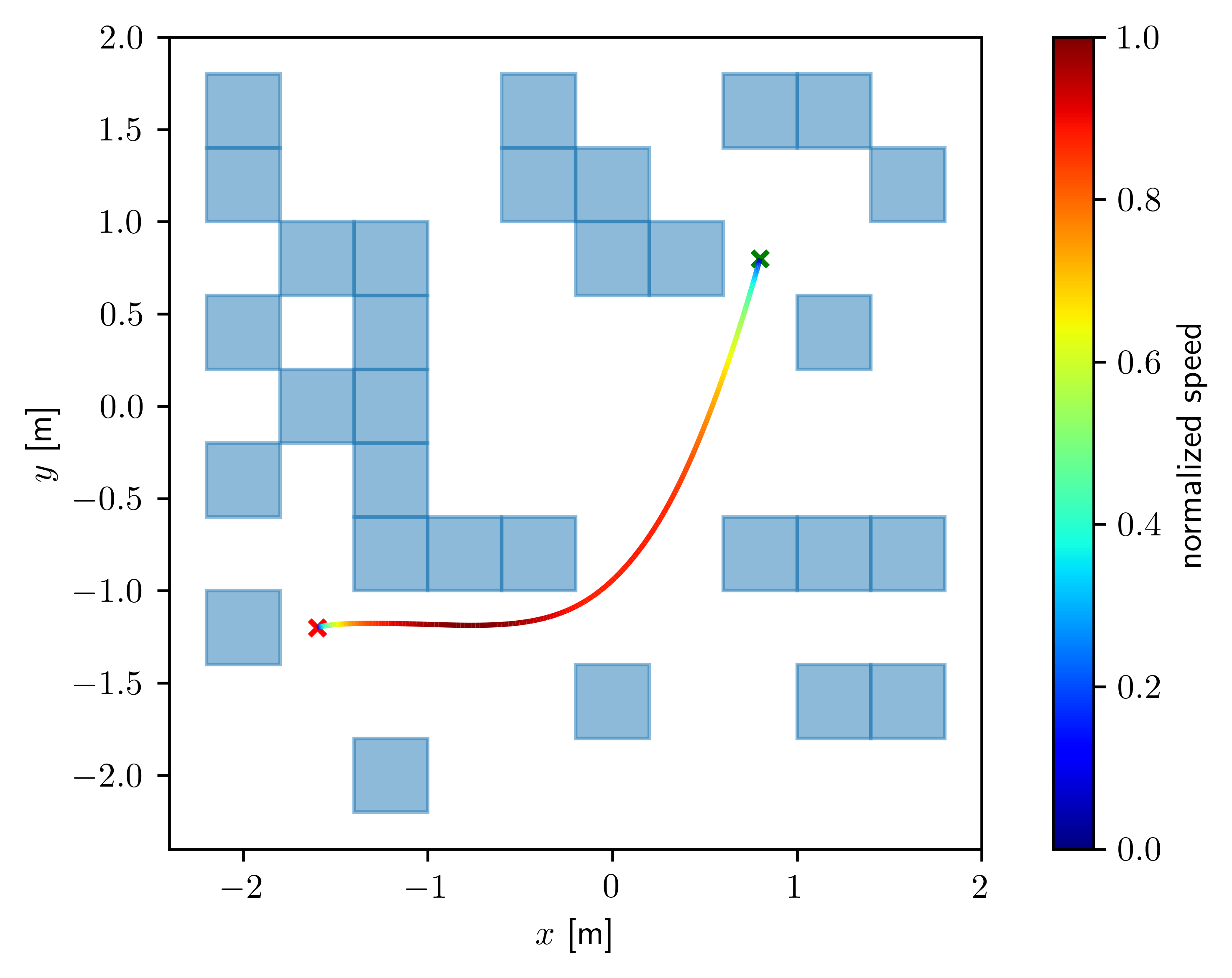}
        \caption{single-seg. a. rate init. with control and forced constraints}
        \label{fig:traj-random-columns-arate-ctrl}
    \end{subfigure}
    \vfill
    \begin{subfigure}[t]{0.45\linewidth}
        \centering
        \includegraphics[width=\linewidth]{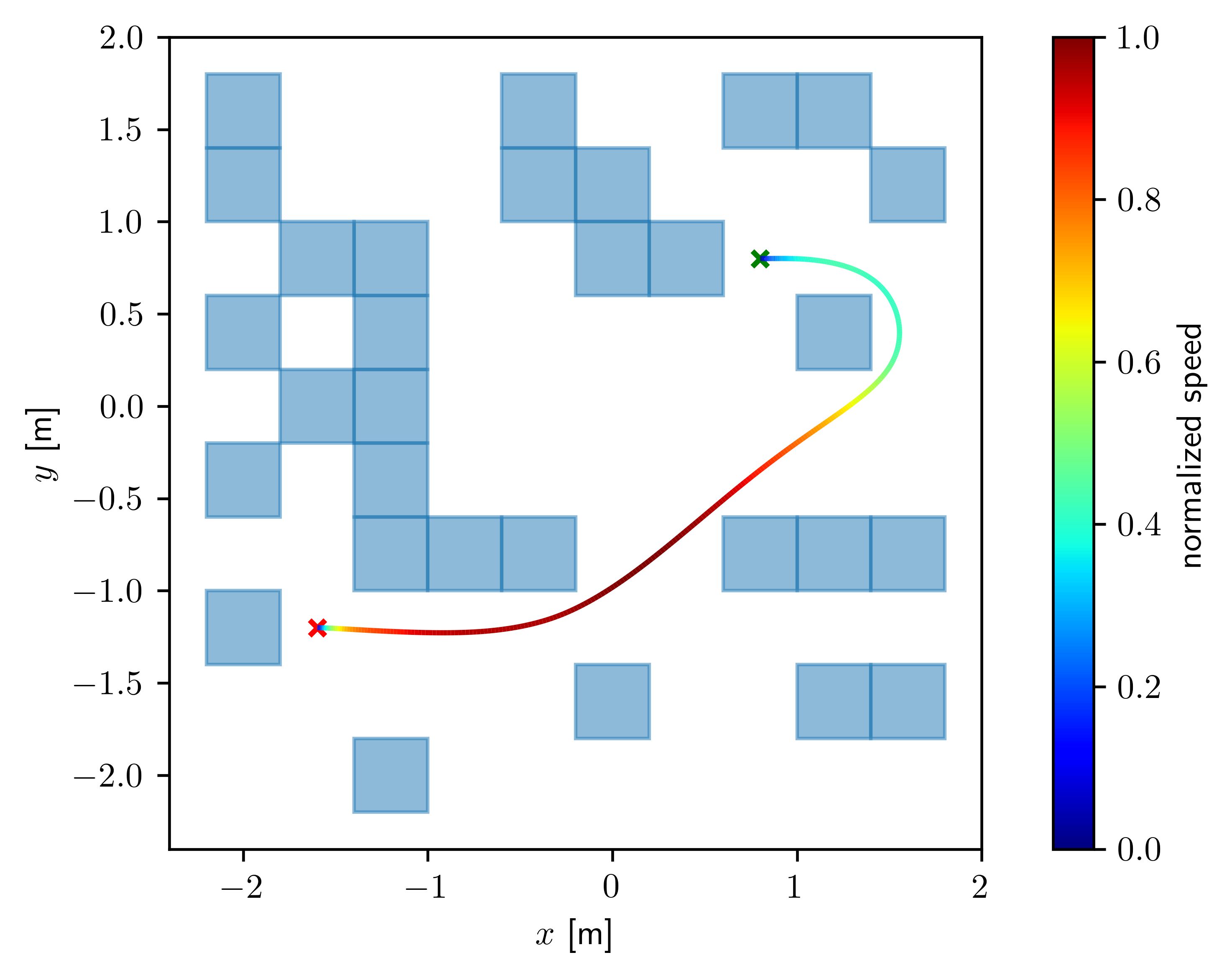}
        \caption{multi-segment simple init.}
        \label{fig:traj-random-columns-none-multi}
    \end{subfigure}
    \hfill
    \begin{subfigure}[t]{0.45\linewidth}
        \centering
        \includegraphics[width=\linewidth]{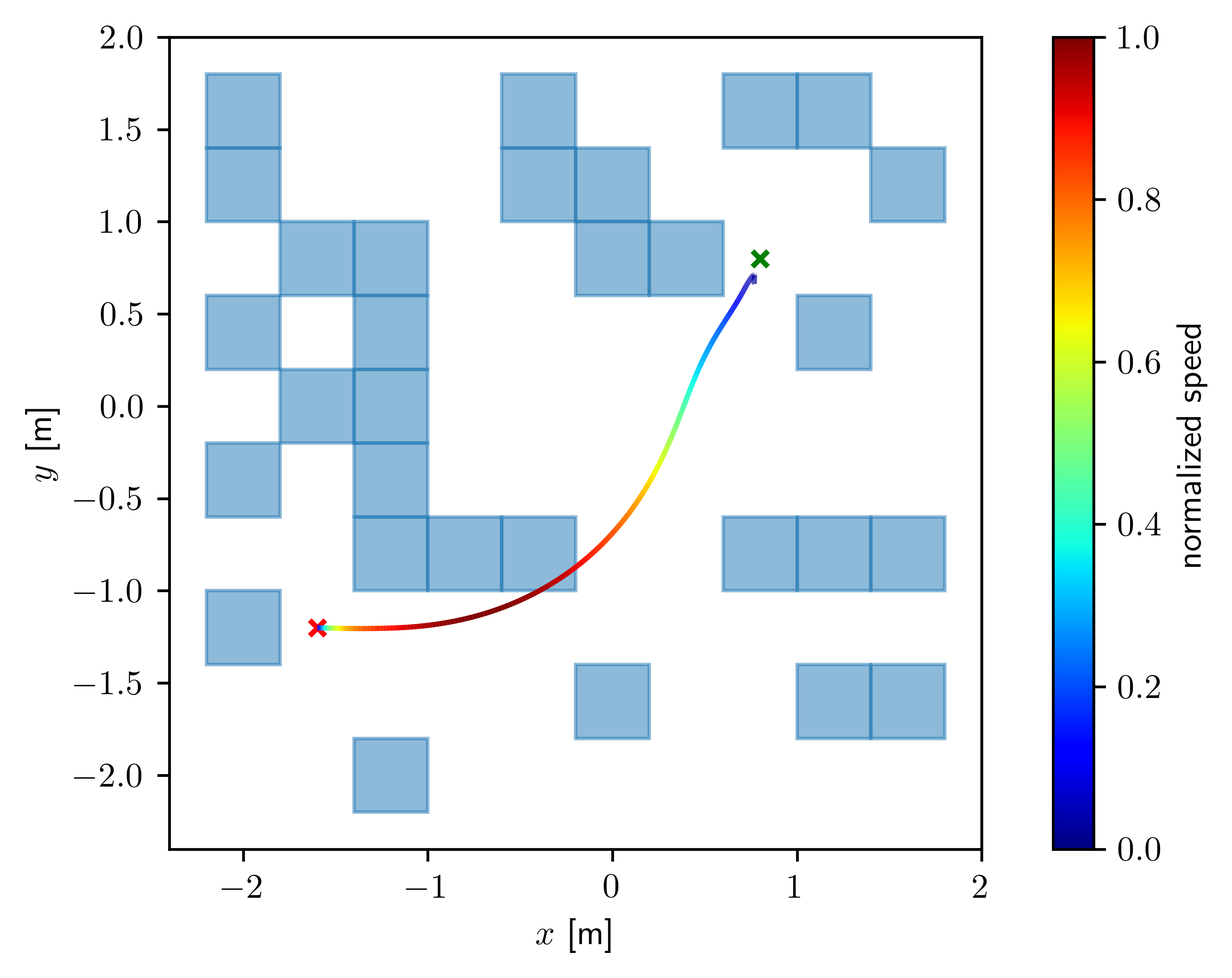}
        \caption{multi-segment position init.}
        \label{fig:traj-random-columns-position}
    \end{subfigure}
    \caption{PSEM trajectory planning for the environment with randomly generated columns}
    \label{fig:traj-plan-random-columns}
\end{figure}

The random columns scenario (\Cref{fig:traj-plan-random-columns}) with 30 obstacles provided diverse insights. Unlike in the two-obstacle scenario, trajectories were found for simple initial guesses in both multi and single-segment initializations.

\Cref{tab:traj-plan-eval-random_columns-single} reveals PSEM with simple initial guess (\Cref{fig:traj-random-columns-none-single}) found a trajectory in just two iterations. PSEM with angular rate and control initialization (\Cref{fig:traj-random-columns-arate-ctrl}) achieved the best maximum absolute error in three iterations.

\input{tables/table_traj_plan_eval_random_columns_single}

For multi-segment initialization (\Cref{tab:traj-plan-eval-random_columns-multi}), simple initialization had the lowest constraint violation but the highest computation time. Position initialization (\Cref{fig:traj-random-columns-position}) was most efficient, found in one iteration with the best criterion.

\input{tables/table_traj_plan_eval_random_columns_multi}

criterion values varied more significantly in this scenario, with no clear advantage between single and multi-segment initialization, suggesting the optimal choice depends on specific scenario characteristics.





\section{Conclusion}
\label{sec:conclusion}



The influence of LT*-based initial guesses on the UAV trajectory planning was investigated using PSM and PSEM. The study revealed that PSEM generally requires fewer iterations but is often computationally more demanding than PSM due to the greater number of collocation points. Our findings indicate that the quality of the initial guess significantly affects the solution. In sparse environments, simple initial guesses often fail, whereas in complex scenarios, they occasionally succeed. Multi-segment position initial guesses with PSEM are computationally efficient and yield low values of the criterion. However, they are prone to constraint violations and collisions.

We suggest that future work should focus on improving collision avoidance by incorporating collision rates and constraint violations into stop conditions, optimizing segmentation strategies, and enhancing dynamics-based initial guesses.

\section*{ACKNOWLEDGMENT}
\small
This work was supported by the Technology Agency of the Czech Republic, programme National Competence Centres, project \#TN 0200 0054 Bozek Vehicle Engineering National Competence Center. Computational resources were provided by the e-INFRA CZ project (ID:90254), supported by the Ministry of Education, Youth and Sports of the Czech Republic.

\bibliographystyle{IEEEtran}
\bibliography{mybib}

\end{document}

%% file: symbols.tex
\newcommand{\abs}[1]{\lvert#1\rvert}
\newcommand{\norm}[1]{\lVert#1\rVert}

\newcommand{\tp}{^\mathsf{T}}

\newcommand{\bdq}{\boldsymbol{q}}
\newcommand{\bdr}{\boldsymbol{r}}
\newcommand{\bds}{\boldsymbol{s}}

\newcommand{\bdx}{\boldsymbol{x}}

\newcommand{\bdu}{\boldsymbol{u}}

\newcommand{\bdz}{\boldsymbol{z}}

\newcommand{\bdD}{\boldsymbol{D}}

\newcommand{\bdF}{\boldsymbol{F}}

\newcommand{\bdI}{\boldsymbol{I}}

\newcommand{\bdK}{\boldsymbol{K}}

\newcommand{\bdQ}{\boldsymbol{Q}}
\newcommand{\bdR}{\boldsymbol{R}}
\newcommand{\bdS}{\boldsymbol{S}}

\newcommand{\bdGamma}{\boldsymbol{\Gamma}}

\newcommand{\bdomega}{\boldsymbol{\omega}}

\newcommand{\bdtau}{\boldsymbol{\tau}}

\newcommand{\x}{x}
\newcommand{\y}{y}
\newcommand{\z}{z}
\newcommand{\localframe}{L}
\newcommand{\bodyframe}{B}
\newcommand{\position}{\bdr}
\newcommand{\vecx}{\vec{\x}}
\newcommand{\vecy}{\vec{\y}}
\newcommand{\vecz}{\vec{\z}}
\newcommand{\state}{\bdx}
\newcommand{\control}{\bdu}
\newcommand{\quaternion}{\bdq}
\newcommand{\arate}{\bdomega}
\newcommand{\force}{\bdF}
\newcommand{\collectivethrust}{\force_T}
\newcommand{\forceaerodyn}{\force_A}
\newcommand{\forceBdes}{\force^\bodyframe_r}
\newcommand{\forceLdes}{\force^\localframe_r}
\newcommand{\quatdynmatrix}{\bdGamma}
\newcommand{\torque}{\bdtau}

%% file: tables/table_traj_plan_eval_simple2_multi.tex
\begin{table}
\small

\caption{Evaluation of UAV trajectories found by PSM and PSEM for 2 obstacles with multi-segment initialization.}
\label{tab:traj-plan-eval-simple2-multi}
\resizebox*{\linewidth}{!}{
\begin{tabular}{p{15mm}rrrp{12mm}p{12mm}rp{12mm}r}
\toprule
Init. Level & Constr. & Method & Iter. & Optimality Criterion & Absolute Error & Sum Viol. & Obstacle Viol. & Total Time \\
\midrule
position & yes & PSEM & \cellcolor{color3} 1 & 2.72e+01 & 4.20e-03 & 2.00e-02 & 1.99e-02 & \cellcolor{color3} 80.94s \\
velocity & yes & PSEM & 5 & \cellcolor{color8} 2.97e+01 & \cellcolor{color8} 5.09e-03 & \cellcolor{color8} 2.91e-02 & \cellcolor{color8} 2.91e-02 & 2308.56s \\
orientation & yes & PSEM & \cellcolor{color8} 7 & \cellcolor{color3} 2.71e+01 & \cellcolor{color3} 3.75e-03 & 1.71e-02 & 1.61e-02 & \cellcolor{color8} 5547.63s \\
a. rate & yes & PSEM & 5 & \cellcolor{color8} 2.97e+01 & \cellcolor{color8} 5.09e-03 & \cellcolor{color8} 2.91e-02 & \cellcolor{color8} 2.91e-02 & 2003.10s \\
a. rate ctrl & yes & PSEM & 3 & 2.82e+01 & 3.81e-03 & \cellcolor{color3} 1.46e-02 & \cellcolor{color3} 1.37e-02 & 972.14s \\
\bottomrule
\end{tabular}}
\end{table}

%% file: tables/table_traj_plan_eval_simple2_single.tex
\begin{table}
\small

\caption{Evaluation of UAV trajectories found by PSM and PSEM for 2 obstacles with single-segment initialization.}
\label{tab:traj-plan-eval-simple2-single}
\resizebox*{\linewidth}{!}{
\begin{tabular}{p{15mm}rrrp{12mm}p{12mm}rp{12mm}r}
\toprule
Init. Level & Constr. & Method & Iter. & Optimality Criterion & Absolute Error & Sum Viol. & Obstacle Viol. & Total Time \\
\midrule
position & yes & PSEM & \cellcolor{color3} 3 & 3.10e+01 & 7.78e-03 & \cellcolor{color8} 2.41e-02 & \cellcolor{color8} 2.11e-02 & 167.66s \\
position & yes & PSM & \cellcolor{color8} 8 & \cellcolor{color8} 3.17e+01 & 9.16e-03 & \cellcolor{color10} 5.62e-03 & \cellcolor{color3} 3.30e-03 & 241.98s \\
velocity & yes & PSM & 6 & \cellcolor{color10} 2.89e+01 & \cellcolor{color8} 9.62e-03 & 9.03e-03 & 6.51e-03 & \cellcolor{color10} 105.95s \\
velocity & yes & PSEM & 7 & 3.00e+01 & \cellcolor{color10} 4.37e-03 & 1.34e-02 & 1.17e-02 & \cellcolor{color9} 730.64s \\
orientation & yes & PSM & \cellcolor{color8} 8 & \cellcolor{color3} 2.88e+01 & \cellcolor{color3} 3.43e-03 & 5.74e-03 & 3.62e-03 & 163.24s \\
angular rate & yes & PSM & 6 & \cellcolor{color10} 2.89e+01 & \cellcolor{color8} 9.62e-03 & 9.03e-03 & 6.51e-03 & 123.88s \\
angular rate & yes & PSEM & 7 & 3.00e+01 & \cellcolor{color10} 4.37e-03 & 1.34e-02 & 1.17e-02 & 476.32s \\
a. rate & yes & PSM & 5 & 2.93e+01 & 8.66e-03 & 7.91e-03 & 6.20e-03 & \cellcolor{color3} 97.57s \\
a. rate ctrl & yes & PSEM & \cellcolor{color8} 8 & 3.03e+01 & 7.87e-03 & \cellcolor{color3} 4.71e-03 & 3.37e-03 & \cellcolor{color8} 1005.08s \\
\bottomrule
\end{tabular}}
\end{table}

%% file: tables/table_traj_plan_eval_random_columns_single.tex
\begin{table}
\small

\caption{Evaluation of UAV trajectories found by PSM and PSEM for random columns with single-segment init.}
\label{tab:traj-plan-eval-random_columns-single}
\resizebox*{\linewidth}{!}{
\begin{tabular}{p{15mm}rrrp{12mm}p{12mm}rp{12mm}r}
\toprule
Init. Level & Constr. & Method & Iter. & Optimality Criterion & Absolute Error & Sum Viol. & Obstacle Viol. & Total Time \\
\midrule
none & no & PSEM & \cellcolor{color3} 2 & \cellcolor{color9} 2.10e+01 & 5.38e-03 & 2.83e-03 & 2.01e-03 & 150.63s \\
none & no & PSM & \cellcolor{color8} 5 & 1.86e+01 & \cellcolor{color8} 9.62e-03 & 3.88e-03 & 2.18e-03 & 140.36s \\
position & yes & PSEM & 4 & 1.77e+01 & 4.38e-03 & \cellcolor{color3} 5.98e-04 & \cellcolor{color3} 3.38e-04 & 151.39s \\
position & yes & PSM & \cellcolor{color8} 5 & 1.90e+01 & 8.22e-03 & 4.17e-03 & 2.20e-03 & 154.93s \\
velocity & yes & PSEM & 4 & 1.72e+01 & 1.97e-03 & \cellcolor{color8} 1.04e-02 & \cellcolor{color8} 5.71e-03 & \cellcolor{color9} 394.50s \\
velocity & yes & PSM & \cellcolor{color8} 5 & 1.76e+01 & 7.63e-03 & 2.94e-03 & 2.94e-03 & 101.50s \\
orientation & yes & PSEM & \cellcolor{color10} 3 & \cellcolor{color3} 1.60e+01 & \cellcolor{color10} 1.56e-03 & 6.10e-03 & 3.49e-03 & \cellcolor{color8} 503.45s \\
orientation & yes & PSM & 4 & 1.75e+01 & 3.94e-03 & 4.11e-03 & 4.10e-03 & \cellcolor{color10} 76.55s \\
a. rate & yes & PSEM & 4 & 1.72e+01 & 1.97e-03 & \cellcolor{color8} 1.04e-02 & \cellcolor{color8} 5.71e-03 & 325.69s \\
a. rate & yes & PSM & \cellcolor{color8} 5 & 1.76e+01 & 7.63e-03 & 2.94e-03 & 2.94e-03 & 128.18s \\
a. rate ctrl & yes & PSM & \cellcolor{color10} 3 & \cellcolor{color8} 2.39e+01 & \cellcolor{color9} 8.28e-03 & 3.73e-03 & 3.65e-03 & \cellcolor{color3} 46.71s \\
a. rate ctrl & yes & PSEM & \cellcolor{color10} 3 & \cellcolor{color10} 1.70e+01 & \cellcolor{color3} 1.17e-03 & 2.83e-03 & 1.04e-03 & 299.46s \\
\bottomrule
\end{tabular}}
\end{table}

%% file: tables/table_traj_plan_eval_random_columns_multi.tex
\begin{table}
\small

\caption{Evaluation of UAV trajectories found by PSM and PSEM for random columns with multi-segment init.}
\label{tab:traj-plan-eval-random_columns-multi}
\resizebox*{\linewidth}{!}{
\begin{tabular}{p{15mm}rrrp{12mm}p{12mm}rp{12mm}r}
\toprule
Init. Level & Constr. & Method & Iter. & Optimality Criterion & Absolute Error & Sum Viol. & Obstacle Viol. & Total Time \\
\midrule
none & no & PSEM & \cellcolor{color8} 4 & 2.26e+01 & 7.44e-03 & \cellcolor{color3} 2.10e-04 & \cellcolor{color3} 1.29e-04 & \cellcolor{color8} 1576.18s \\
position & yes & PSEM & \cellcolor{color3} 1 & \cellcolor{color3} 1.49e+01 & \cellcolor{color3} 2.96e-03 & \cellcolor{color8} 3.90e-02 & \cellcolor{color8} 3.89e-02 & \cellcolor{color3} 37.94s \\
velocity & yes & PSEM & 2 & \cellcolor{color8} 2.45e+01 & \cellcolor{color8} 9.37e-03 & 1.49e-02 & 9.26e-03 & 345.94s \\
orientation & yes & PSEM & 3 & 1.62e+01 & 6.39e-03 & 8.81e-04 & 7.96e-04 & 311.26s \\
a. rate & yes & PSEM & 2 & \cellcolor{color8} 2.45e+01 & \cellcolor{color8} 9.37e-03 & 1.49e-02 & 9.26e-03 & 338.72s \\
a. rate ctrl & yes & PSEM & 3 & 1.76e+01 & 4.66e-03 & 1.53e-02 & 1.01e-02 & 346.07s \\
\bottomrule
\end{tabular}}
\end{table}